\documentclass{article}

     \usepackage[preprint,nonatbib]{neurips_2019}

\usepackage[utf8]{inputenc} 
\usepackage[T1]{fontenc}    
\usepackage{hyperref}       
\usepackage{url}            
\usepackage{booktabs}       
\usepackage{amsfonts}       
\usepackage{nicefrac}       
\usepackage{microtype}      
\usepackage{subfigure}
\usepackage{amsmath}
\usepackage{amssymb}
\usepackage{amsthm}
\usepackage{mathtools}
\usepackage{stfloats}
\usepackage{multirow}
\usepackage[toc,page]{appendix}

\newcommand{\eg}{\emph{e.g. }}
\newcommand{\etal}{\emph{et.al.}}
\newcommand{\ie}{\emph{i.e. }}
\newcommand{\etc}{\emph{etc}}
\newcommand{\wrt}{\emph{w.r.t. }}

\newcommand{\vect}{\text{vec}}

\usepackage{hyperref}
\usepackage{algorithm}
\usepackage{algpseudocode}

\title{Full-Stack Filters to Build Minimum Viable CNNs}

\author{%
  Kai Han$^1$, Yunhe Wang$^1$, Yixing Xu$^1$, Chunjing Xu$^1$, Dacheng Tao$^2$, Chang Xu$^2$\\
  $^1$Huawei Noah's Ark Lab\\
  $^2$School of Computer Science, FEIT, University of Sydney, Australia\\
  \texttt{\{kai.han, yunhe.wang, yixing.xu, xuchunjing\}@huawei.com}\\
  \texttt{\{dacheng.tao,c.xu\}@sydney.edu.au} \\
}

\begin{document}

\maketitle

\begin{abstract}
	Deep convolutional neural networks (CNNs) are usually over-parameterized, which cannot be easily deployed on edge devices such as mobile phones and smart cameras. Existing works used to decrease the number or size of requested convolution filters for a minimum viable CNN on edge devices. In contrast, this paper introduces filters that are full-stack and can be used to generate many more sub-filters. Weights of these sub-filters are inherited from full-stack filters with the help of different binary masks. Orthogonal constraints are applied over binary masks to decrease their correlation and promote the diversity of generated sub-filters. To preserve the same volume of output feature maps, we can naturally reduce the number of established filters by only maintaining a few full-stack filters and a set of binary masks. We also conduct theoretical analysis on the memory cost and an efficient implementation is introduced for the convolution of the proposed filters. Experiments on several benchmark datasets and CNN models demonstrate that the proposed method is able to construct minimum viable convolution networks of comparable performance.
\end{abstract}

\section{Introduction}
Convolutional neural networks (CNNs) have shown extraordinary success in a large variety of computer vision tasks, such as image recognition~\cite{Simonyan15,he2016deep}, object detection~\cite{renNIPS15fasterrcnn}, and semantic segmentation~\cite{long2015fully}. For the accuracy reason, most of widely used CNNs are heavily designed and running on GPU servers and clusters, \eg over 500 \emph{MB} memory are required by the VGG-16~\cite{Simonyan15} for storing its convolution filters and more than $1\times10^{10}$ FLOPs are needed for processing an image of standard size. However, CNNs running on edge devices should be of \textit{minimum} size yet still \textit{viable} to reach satisfactory performance.

For compatibility with edge devices, a number of methods have been developed to slim down existing heavy CNNs. For instance, Han \etal~\cite{han16} and Zhou \etal~\cite{zhou2016less} investigated the sparsity of convolution filters and pruned redundant weights in pre-trained deep models. Wang \etal~\cite{wang2016cnnpack} further addressed the problem in the DCT frequency domain, which shows that over $75\%$ of weights in pre-trained networks (\eg ResNet-50 and VGG-16) can be discarded without any accuracy drop. Besides the pruning scheme, there are considerable methods for learning a portable alternative to the original heavy network such as quantization and binarization~\cite{arora2014provable,gupta2015deep,rastegari2016xnor}, matrix and tensor decomposition~\cite{denton2014exploiting,jaderberg2014speeding,tai2016convolutional,savarese2018learning}, and knowledge distillation paradigm~\cite{Distill,romero2014fitnets,balan2015bayesian}. 

Beyond manipulating (\eg pruning and compressing) existing over-parameterized neural networks, there are methods to explore lightweight architectures specially designed for mobile platforms.
For instance, He \etal~\cite{he2016deep} introduced shortcut connections and replaced most of $3\times3$ filters by $1\times1$ filters. Shang \etal~\cite{shang2016understanding} learned filters with pair-wise positive-negative phase constraint. Cohen \etal~\cite{cohen2016group} utilized translation, reflection and rotation transformations for augmenting filters. Howard \etal~\cite{howard2017mobilenets} proposed a MobileNet for portable devices with depth-wise convolutions and massive $1\times1$ convolution filters. Zhang \etal~\cite{zhang2018shufflenet} exploited group convolution and channel shuffle operation to construct compact CNNs with higher performance. Wang \etal~\cite{Versatile} proposed a series of versatile filters to establish portable CNNs. Sandler \etal~\cite{mobilev2} and Ma \etal~\cite{shufflev2} further designed some effective modules to enhance the performance of these compact CNN models.



These aforementioned methods have produced impressive efficient CNNs, either by decreasing the number of filters (weights) or advocating smaller filters (\eg $1\times 1$). For example, parameters of a well-trained MobileNet~\cite{howard2017mobilenets} with a $29.4\%$ top-1 error only occupy 16 MB and 569 MFLOPs are requested to process an image of $224\times 224$. An immediate question naturally arises: how much further can we compress and accelerate CNNs while preserving their accuracy? Instead of following classical solutions to decrease the established computational load, we provide a new way of thinking that starts with some basic convolutions first and then gradually introduces more computations to approach minimum viable CNNs. 

In this paper, we develop full-stack filers to build minimum viable CNNs. ``Full stack'' refers to the collection of a series of characterizes needed for a filter to extract useful information from the input. We employ binary masks to distill a full-stack filter down to distinct sub-filters for different convolutions. Orthogonal constraints are exploited to promote the diversity of these sub-filters. Both full-stack filters and binary masks are learned in an end-to-end manner through a specifically designed optimization method. The resulting neural network is lightweight, as only a few full-stack filters are in 32 floating point resolution while associated masks are in 1-bit format.  Memory cost is discussed in detail and an efficient method is provided to compute the convolution results of sub-filers by investigating their connection with the full-stack filters. Experiments on benchmark datasets and models demonstrate that the proposed convolution operation can be beneficial for establishing extremely compact CNNs with comparable performance.

\section{Minimum Viable CNNs}
We aim to develop CNNs that have minimum size  yet still viable to reach an accuracy as high as possible. A novel full-stack filter is designed to decrease the computational cost requested for feature extraction.  

\subsection{Full-stack Filters}
For a given arbitrary convolutional layer, we denote the input data of this layer as $X\in\mathbb{R}^{c\times h\times w}$, the output as $Y\in\mathbb{R}^{n\times h'\times w'}$, and the convolution filters as $F\in\mathbb{R}^{n\times c\times d\times d}$, where $c$ is the number of channels in the input data, $n$ is the number of convolution filters in this layer, \ie the number of channels in the output data, $h$, $w$, $h'$, and $w'$ are heights and widths of the input data and output data, respectively. In addition, $d\times d$ is the size of convolution filters. Here we only discuss square filters for simplicity and the proposed method can be naturally applied to filters of any sizes. The conventional convolution can be formulated as
\begin{equation}\label{eq:conv}
[Y_1,...,Y_n]  = [F_1*X,...,F_n*X],
\end{equation}
where $*$ denotes the convolution operation in CNNs, $Y_n\in\mathbb{R}^{h'\times w'}$ is the $n$-th feature map of $Y$ through the $n$-th convolution filter $F_n\in\mathbb{R}^{c\times d\times d}$. 

Dividing the input $X$ into $l=h'\times w'$ patches $\{x_1,\cdots,x_l\}$ where $x_l\in\mathbb{R}^{c\times d\times d}$ by sliding window, we vectorize them and have $\mathbf{X}=[\vect(x_1),\cdots,\vect(x_l)]\in\mathbb{R}^{(cd^2)\times l}$, where $\vect(\cdot)$ is the vectorization operation for converting matrices or tensors to vectors. Similarly, the output feature map and the convolutional filters can be reformulated as $\mathbf{Y}=[\vect(Y_1),\cdots,\vect(Y_n)]\in\mathbb{R}^{l\times n}$ and $\mathbf{F}=[\vect(F_1),\cdots,\vect(F_n)]=[\mathbf{f}_1,\cdots,\mathbf{f}_n]\in\mathbb{R}^{(cd^2)\times n}$, respectively. Then the convolution operation is rewritten as
\begin{equation}\label{eq:conv2}
\mathbf{Y} = \mathbf{X}^T\mathbf{F} = [\mathbf{X}^T\mathbf{f}_1,...,\mathbf{X}^T\mathbf{f}_n],
\end{equation}
where the $i$-th column in $\mathbf{Y}$ is the convolution result of all the patches using the $i$-th filter in $\mathbf{F}$. 
Compared with traditional filters in fully-connected neural networks, the biggest advantage of convolution filters is sharing weights to different spatial locations. For example, a $3\times3$ convolution filter will scan each $3\times3$ area on the input data with a size of $224\times224$, while the size of filters in a fully-connected layer have to be equal to $224\times224$.

Each convolution response in feature maps used to be calculated through hundreds or thousands of parameters in a convolution filter. Although smaller convolution filters are increasingly popular, \eg massive $1\times1$ filters are used in MobileNet~\cite{howard2017mobilenets} and ShuffleNet~\cite{zhang2018shufflenet}, the number of requested filters is often large to preserve the capacity of the layer.  To maximize the value of limited filters, we propose to learn a series of full-stack filters, which can take more roles than classical filters. The strengths of these full-stack filters can be fully disclosed by applying different binary masks.

\begin{figure*}[t]
	\centering
	\begin{tabular}{cc}
		\includegraphics[width=0.45\columnwidth]{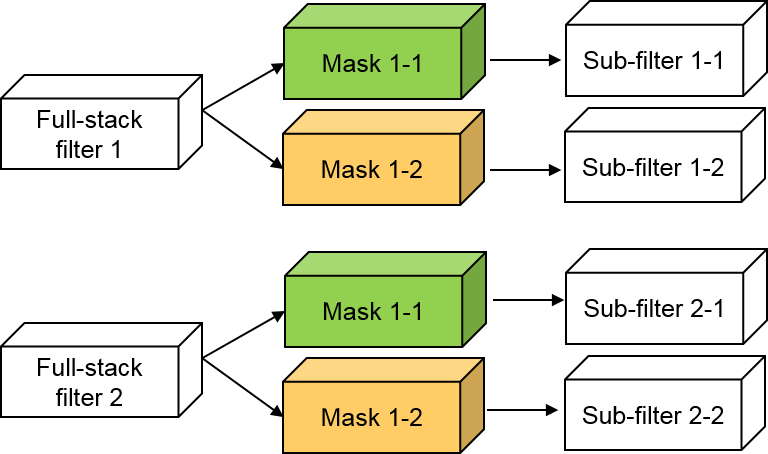} \hspace{0.2cm} &\includegraphics[width=0.45\columnwidth]{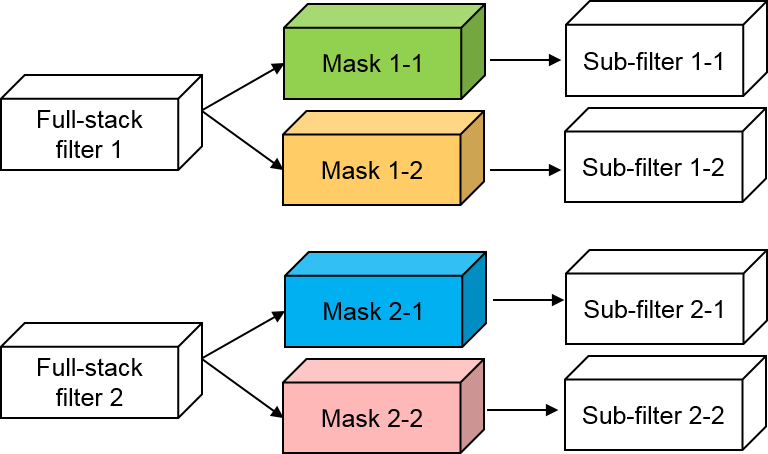}
		\\
		\small (a) Sub-filters generation using shared masks. & \small (b) Sub-filters generation using individual masks.
	\end{tabular}
	\caption{An illustration of the proposed method for generating sub-filters using full-stack filters and different mask strategies, \ie shared and individual associated masks, which can be selected according to the practical requirement. Wherein, associated masks in the same color are exactly the same one. Better view in the color version.} 
	\label{Fig:illustration}
	\vspace{-1em}
\end{figure*}

Given a full-stack filter $\mathbf{b}\in\mathbb{R}^{cd^2}$ and a binary mask $\mathbf{m}\in\{-1,+1\}^{cd^2}$, a sub-filter can be generated through $\mathbf{b}\circ \mathbf{m}$, where $\circ$ is the element wise product. If there are $s$ different binary masks, we can then generate $s$ sub-filters from a full-stack filter $\mathbf{b}$. Suppose that we learn $k$ full-stack filters and  $s$ binary masks to generate the similar number of convolution filters ($k\times s \approx n$) as the original neural network. Thus, Eq.~\ref{eq:conv2} can be reformulated as
\begin{equation}\label{eq:1tor}
\mathbf{Y}=[\mathbf{X}^T\hat{\mathbf{f}}_{11},\mathbf{X}^T\hat{\mathbf{f}}_{12},...,\mathbf{X}^T\hat{\mathbf{f}}_{ks}]= [\mathbf{X}^T(\mathbf{b}_1\circ\mathbf{m}_1),\mathbf{X}^T(\mathbf{b}_1\circ\mathbf{m}_2),...,\mathbf{X}^T(\mathbf{b}_k\circ\mathbf{m}_s)],
\end{equation}
where $\hat{\mathbf{f}}_{ks}$ is the generated sub-filter from the $k$-th full-stack filter and the $s$-th binary mask.

A full-stack filter will be exploited for many times (up to $s$) in the proposed method, but only one copy has to be maintained, which implies that fewer 32-bit values are required than that of standard convolution filters. Meanwhile, computational complexity can be largely reduced by exploiting intermediate convolution results of full-stack filters on the input data. More in-depth discussions are provided in Section \ref{sec:complexity}.

\subsection{Deployment of Binary Masks}
In Eq.~\ref{eq:1tor}, binary masks are shared with different full-stack filters, \eg $\hat{\mathbf f}_{11}$ and $\hat{\mathbf f}_{21}$ are sub-filters generated by applying mask $M_1$ on the first and the second full-stack filters, as shown in Figure~\ref{Fig:illustration}(a). We thus name this kind of deployment of binary masks as the \textit{shared} strategy.

Masks in $\mathbf{M}$ are binary and would not account for an obvious proportion of the entire memory usage of CNNs by the proposed method. It is therefore applicable to assign each full-stack convolution filter with a separate set of $s$ binary masks, so that we can harvest abundant diverse sub-filters at a  small memory cost. To this end, we further extend Eq. \ref{eq:1tor} to contain $k\times s$ binary masks, and each sub-filter is represented as
\begin{equation}\label{eq:ind}
\hat{\mathbf{f}}_{ij}=\mathbf{b}_i\circ \mathbf{m}_{ij},
\end{equation}
where $i=1,\cdots,k$, and $j=1,\cdots,s$. The convolution operation using separate mask sets can be formulated as
\begin{equation}\label{eq:conv-ind}
\mathbf{Y} = [\mathbf{X}^T\hat{\mathbf{f}}_{11},\mathbf{X}^T\hat{\mathbf{f}}_{12},...,\mathbf{X}^T\hat{\mathbf{f}}_{ks}]= [\mathbf{X}^T(\mathbf{b}_1\circ \mathbf{m}_{11}),\mathbf{X}^T(\mathbf{b}_1\circ \mathbf{m}_{12}),...,\mathbf{X}^T(\mathbf{b}_k\circ \mathbf{m}_{ks})],
\end{equation}
whose computational complexity is the same as that of the shared strategy. Although the number of masks is increased, this \textit{separate} strategy can lead to better representation capability, because of the increased diversity of generated sub-filters. 

To further promote the diversity of these sub-filters and their resulting feature maps, binary masks for a full-stack filter should be different from each other as much as possible. Therefore, we propose to decrease the correlation between binary masks. By vectorizing and stacking all $s$ binary masks into a matrix, we have
\begin{equation}
\mathbf{M} = [\mathbf{m}_1,...,\mathbf{m}_s],
\end{equation}
where $\mathbf{M}$ should be an approximately orthogonal matrix for generating feature maps without obvious correlation. We therefore apply the following regularization on binary masks in each convolutional layer during the training stage:
\begin{equation}
\mathcal{L}_{\text{ortho}} = \frac{1}{2}\|\frac{1}{d^2c}\mathbf{M}^T\mathbf{M}-\mathbf{I}\|_F^2,
\end{equation}
where $\mathbf{I}\in\mathbb{R}^{s\times s}$ is an identity matrix, $\|\cdot\|_F$ is the Frobenius norm for matrices, and $d^2c$ is the number of elements in each column of $\mathbf{M}$. By minimizing the above function, binary masks would be approximately orthogonal with each other, which will improve the difference between sub-filters and increase the informativeness of resulting feature maps.

\subsection{Learning Filters and Masks}

In this section, we introduce how to learn the filters and masks in the developed minimum viable CNNs. Shared and separate strategies are designed to deploy the binary masks for full-stack filters. Here we take the shared strategy as an example to illustrate the optimization.

There are two sets of learnable variables, \ie $\mathbf{B}=[\mathbf{b}_1,...,\mathbf{b}_k]\in\mathbb{R}^{(cd^2)\times k}$ and $\mathbf{M}=[\mathbf{m}_1,...,\mathbf{m}_s]\in{\{-1,+1\}}^{(cd^2)\times s}$. The objective function of the proposed algorithm can be written as
\begin{equation}\label{eq:object}
\min_{\mathbf{B},\mathbf{M}} \mathcal{L} = \mathcal{L}_{0}(\mathbf{B},\mathbf{M}) + \lambda\mathcal{L}_{\text{ortho}}(\mathbf{M}),
\end{equation}
where $\mathcal{L}_{0}$ indicates a loss function for the task at hand, \eg cross entropy loss for classification task and mean square error loss for regression task, and $\lambda$ is the hyper-parameter for orthogonal regularization. The gradients to $\mathbf{B}$ can be calculated via standard back-propagation rule. As for $\mathbf{M}$, we adopt STE (straight-through estimator) in~\cite{bengio2013estimating} to estimate the gradients. Then all the parameters in the entire model are updated by stochastic gradient descent (SGD) optimizer. We call the learned minimum viable CNNs (MVNets) using shared and separate masks as MVNet-A and MVNet-B, respectively. The details of the gradients calculation and the optimization algotihtm can be found in the supplementary materials.

\subsection{Analysis on Complexities}\label{sec:complexity}
Compared with classical CNNs, MVNet can use less parameters to establish the same number of filters in each convolutional layer. In this section, we analyze the memory usage of MVNet in detail. We also provide an efficient implementation for the proposed convolutions, which theoretically brings in fewer multiplication operations.

\paragraph{Model Size Compression}
Given filters $F\in\mathbb{R}^{n\times c\times d\times d}$ in a classical convolution layer, the number of parameters is $n\times c\times d\times d$ 32-bit floating values (omitting bias term). To establish $n$ filters of the size $c\times d\times d$, parameters of full-stack filters in the proposed convolution with shared binary masks are $k\times c\times d\times d$ 32-bit floating values, and the shared masks have $n/k\times c\times d\times d$ 1-bit binary values. The parameter compression ratio can be calculated by
\begin{equation}
r_1 = \frac{n c d^2}{k c d^2 + \frac{1}{32}s c d^2} = \frac{s}{1+\frac{s}{32k}}.
\end{equation}
Given the separate strategy, the number of parameters in full-stack filters is the same, that is, $k\times c\times d\times d$ 32-bit floating values. But separate masks need $n\times c\times d\times d$ 1-bit binary values. Overall, the compression ratio is
\begin{equation}
r_1 = \frac{n c h w}{k c h w + \frac{1}{32}n c h w} = \frac{s}{1+\frac{s}{32}}.
\end{equation}
It can be seen that MVNet-A with shared masks leads to a more compact CNN of fewer parameters, while MVNet-B with separate masks can increase the diversity of sub-filters and bring in richer representations of the input. The effects of both MVNet-A and MVNet-B are thoroughly evaluated in the following experiments.

\paragraph{Speeding Up Convolutions}\label{sec:mul}
In the conventional implementation of convolution operation \cite{jia2014caffe,tensorflow2015-whitepaper}, the input feature map is split into many patches and convolution is applied on each patch to compute the output feature map. Given a patch $x\in\mathbb{R}^{c\times d\times d}$ in the input feature map, the conventional convolution operation can be represented as
\begin{equation}
\begin{aligned}
Y &= \left\{ F_i * x \right\}_{i=1}^{n} = \left\{ \sum\left(\vect(x)\circ\mathbf{f}_i\right) \right\}_{i=1}^{n},
\end{aligned}
\end{equation}
where $Y\in\mathbb{R}^{n\times 1\times 1}$. It involves $n c d^2$ multiplications (MUL) and $n c d^2$ add operations (ADD). In the proposed method, we conduct the convolution as
\begin{equation}\label{eq:old-mask}
Y = \left\{ \sum\left( \vect(x)\circ\hat{\mathbf{f}}_{ij}\right) \right\}_{i=1,j=1}^{k,s}= \left\{ \sum\left(\vect(x)\circ\mathbf{b}_i\circ\mathbf{m}_j \right)  \right\}_{i=1,j=1}^{k,s}.
\end{equation}
It is instructive to note that $\vect(x)\circ \mathbf{b}_i$ has been repeatedly calculated for different masks $\mathbf{m}_j$. We thus only compute it for once and store it in cache. Given the intermediate result $\{\mathbf{c}_i=\vect(x)\circ \mathbf{b}_i\}_{i=1}^{k}$, Eq.~\ref{eq:old-mask} can be simplified as
\begin{equation}
Y = \left\{ \sum\left(\mathbf{c}_i\circ \mathbf{m}_j\right)  \right\}_{i=1,j=1}^{k,s}.
\end{equation}
Noticing that $\mathbf{m}_j$ is binary, multiplication with it can be implemented efficiently by binary masking operation in practice, which takes much less time than standard multiplication. In total, the convolution in our method for one patch only needs $k c d^2$ MULs, $n c d^2$ ADDs and the negligible $n c d^2$ masking operations. Compared to the conventional convolution, the reduction ratio of the number of MUL operations is 
\begin{equation}
r_2 = \frac{n c d^2}{k c d^2} = s.
\end{equation}
In order to maximize the speed-up effect of the multiplication reduction in full-stack filters, more engineering efforts could be taken to further optimize the efficiency of full-stack filters on various devices, e.g. FPGA and NPU. 

\section{Experiments}
In this section, we conduct experiments to validate the effectiveness of the proposed MVNet on benchmark classification datasets: MNIST \cite{lecun1998gradient}, CIFAR-10 \cite{krizhevsky2009learning} and ImageNet \cite{deng2009imagenet}, and the object detection dataset MS COCO \cite{lin2014microsoft}. The results are analyzed for further understanding the superiority of the proposed method. We refer to the minimum viable CNNs using shared and separate masks as MVNet-A and MVNet-B, respectively.

\begin{table}[htbp]
	\centering
	\vspace{-0.5em}
	\caption{The performance of MVNets on MNIST. \#param (32-bit) means the number of parameters counting in 32 bit, MULs means the number of multiplications, and ACC means the accuracy.}\label{tab:mnist}
	\small\vspace{-0.5em}
	\begin{tabular}{c||c|c|c}
		\hline
		{Method}  & \#param (32-bit) & \#MUL (M) & ACC (\%) \\
		\hline\hline
		Original  & $4.31\times 10^5$ & 2.29 & 99.20 \\
		MVNet-A ($s$=10)  & $0.49\times 10^5$ & 0.23 & 99.17 \\
		MVNet-B ($s$=10)  & $0.61\times 10^5$ & 0.23 & 99.21 \\
		MVNet-A ($s$=20)  & $0.28\times 10^5$ & 0.11 & 99.01 \\
		MVNet-B ($s$=20)  & $0.40\times 10^5$ & 0.11 & 99.15 \\
		MVNet-A-Rand ($s$=20)  & $0.28\times 10^5$ & 0.11 & 98.55 \\
		MVNet-B-Rand ($s$=20)  & $0.40\times 10^5$ & 0.11 & 98.57 \\
		\hline
	\end{tabular}
\vspace{-1.0em}
\end{table}

\subsection{Experiments on MNIST}
MNIST dataset consists of 70,000 $28\times28$ gray-scale hand-written digit images in 10 classes (from 0 to 9), which is split into 60,000 training and 10,000 test images.
LeNet \cite{lecun1998gradient} is adopted as the baseline model, which has four convolutional layers of size $20\times1\times5\times5$, $50\times20\times5\times5$, $500\times50\times4\times4$ and $10\times500\times1\times1$, respectively. The baseline model accounts about 1.6MB and has $99.20\%$ accuracy. We replace the conventional convolution filter with our full-stack filters for the first 3 layers, except the last layer. We set $s=10$, and the results in Table \ref{tab:mnist} show that both MVNet-A and MVNet-B can obtain comparable performance with the baseline, meanwhile yielding about $7.1\times$ and $8.8\times$ compression ratios, respectively. Further increasing $s$ to 20 lowers performance slightly but the memory has been reduced significantly and MUL is reduced by $s\times$. On the other side, MVNet-B performs better than MVNet-A consistently. Compared to the shared strategy, the separate strategy can provide more diverse masks and improve the capacity of the network.

To justify the advantages of the learned binary masks, we also test MVNet with random fixed binary masks (MVNet-A-Rand and MVNet-B-Rand in Table \ref{tab:mnist}). The results in Table \ref{tab:mnist} show that the learnable masks enjoy more performance improvement.

\subsubsection{Impact of Hyper-parameters}
\begin{figure*}[t]
	\vspace{-0.5em}
	\centering
	\begin{tabular}{cc}
		\includegraphics[width=0.45\columnwidth]{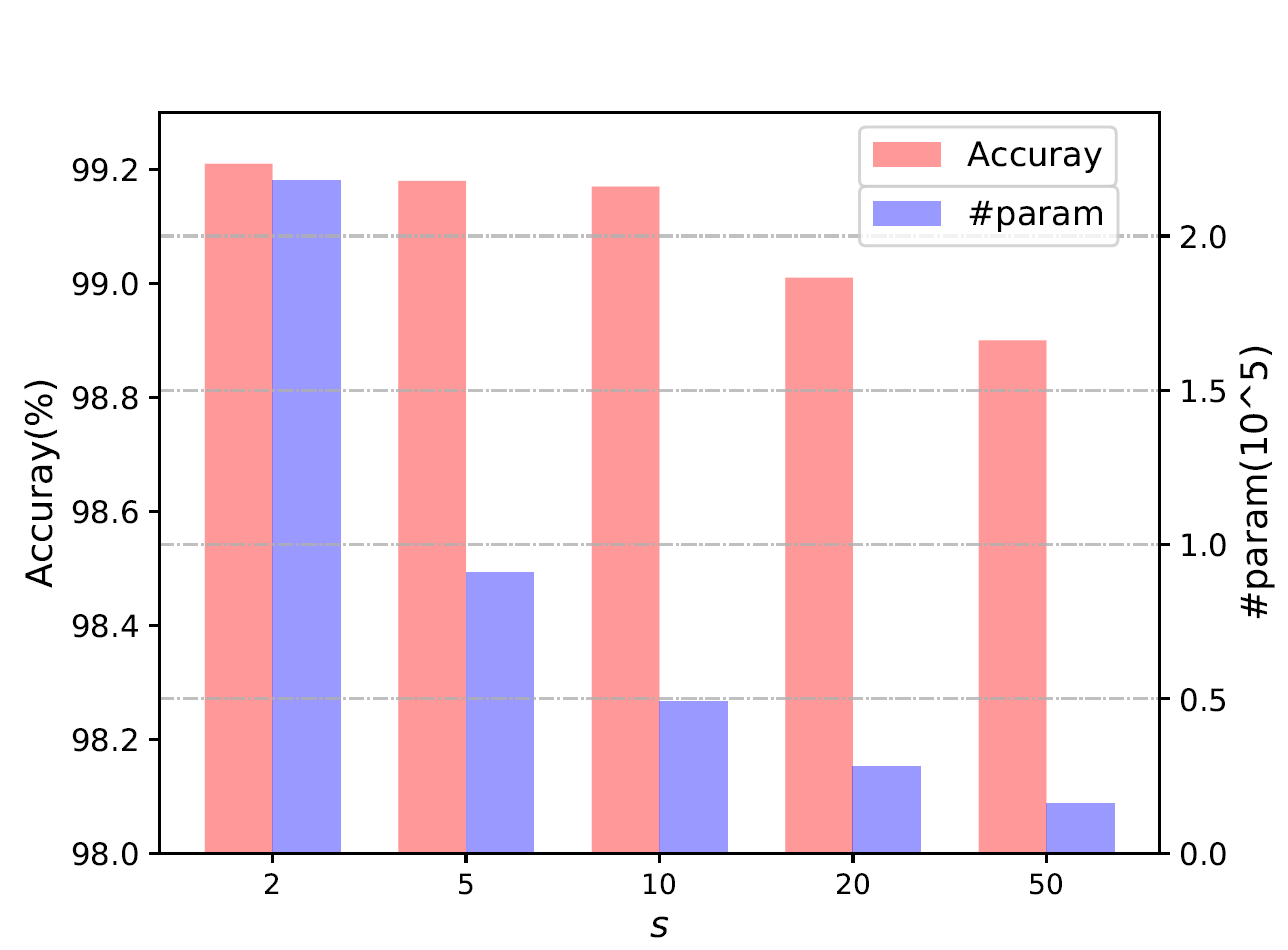} \hspace{0.2cm} &\includegraphics[width=0.45\columnwidth]{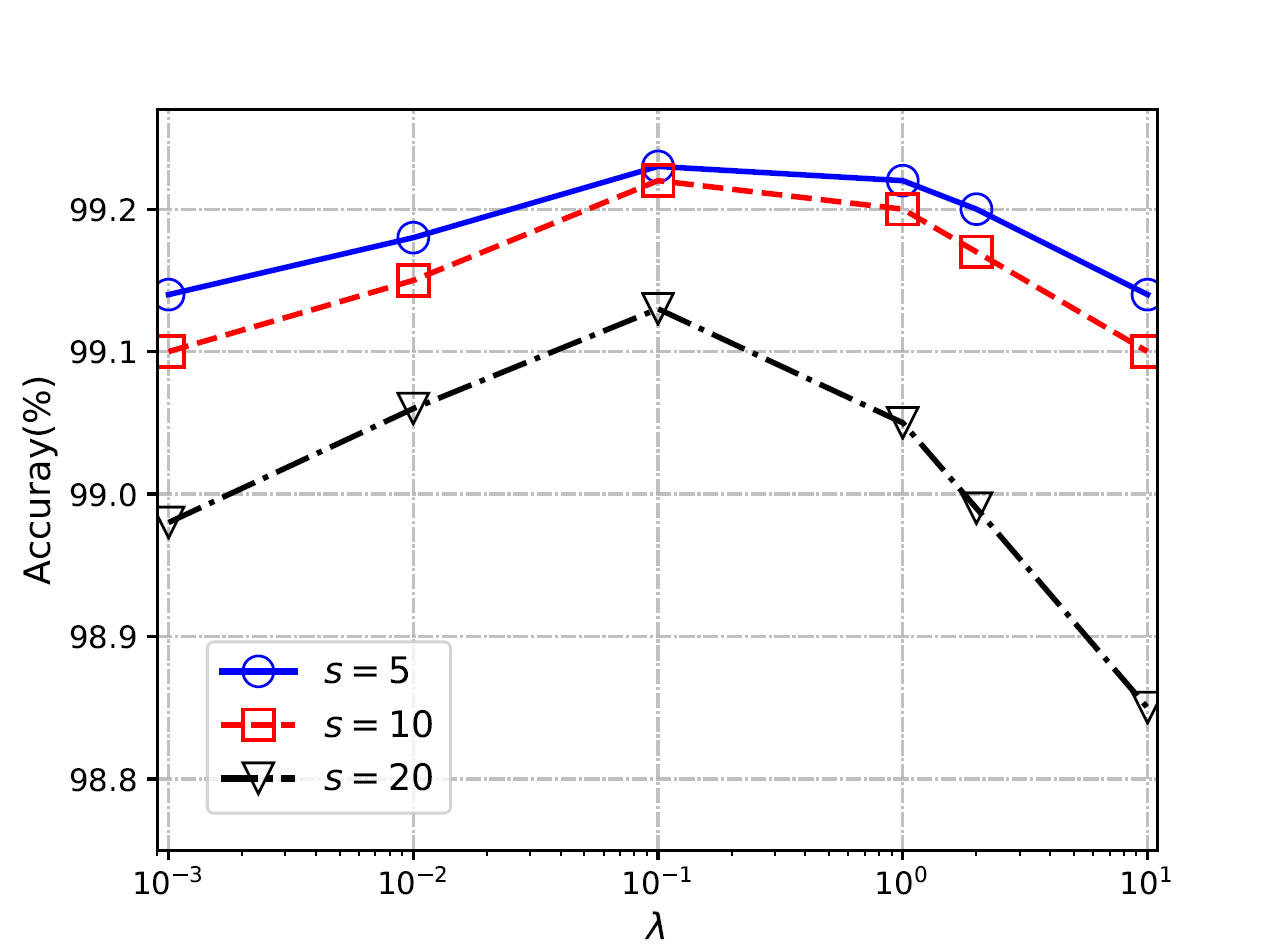}
		\\
		\small (a) The performance of MVNet \wrt different $s$. & \small (b) The performance of MVNet \wrt different $\lambda$.
	\end{tabular}
\vspace{-0.5em}
	\caption{Ablation study of $s$ and $\lambda$. Better view in the color version.} 
	\label{fig:ablation}
	\vspace{-1.5em}
\end{figure*}

The proposed method has shown impressive compression ability while preserving information inside the convolution filters. We further investigate the influence of the compression ratio on the model performance on the MNIST  dataset. LeNet based MVNet-A is adopted here. In practice, $s$ is an important hyper-parameter directly controlling the compression ratio, and in this experiment we set $s$ in the range of $\{2,5,10,20,50\}$. The relation between the number of parameters and the classification accuracy is plotted in Fig. \ref{fig:ablation}(a). With the increase of $s$, the number of model parameters decreases rapidly, but the classification accuracy stays steady until $s$ increases to $20$, and then the accuracy starts to decrease slightly. In the extreme case where $s=50$, the final model only needs $0.16\times10^5$ parameters (compression ratio $>26.9\times$) with a $98.90\%$ classification accuracy. 
Thus our method promises a larger compression ratio meanwhile maintains a higher performance. The value of $s$ can be chosen according to the demand or restrictions of devices.

The orthogonal regularization is used for encouraging diversity of the masks. $\lambda$ is the hyper-parameter for the trade-off between orthogonal regularization and recognition loss. We set $s=\{5,10,20\}$ and tune $\lambda$ from $0.001$ to $10$. The results are reported in Fig.\ref{fig:ablation}(b). For different $s$, the experiment results are consistent with the above observation that smaller $s$ basically achieves better accuracy. The model obtains the best performance around $\lambda=0.1$ (in the following experiments for large-scale datasets, we set $\lambda=0.1$). When $\lambda$ is less than $0.1$, the accuracy decreases gradually; and as $\lambda$ increases from $0.1$, the accuracy stays steady at first and then decreases when $\lambda$ becomes too large to disturb the learning of the recognition loss. The model can obtain a satisfactory performance under a relatively wide range of $\lambda$, so we can choose an appropriate hyper-parameter easily in practice.

\begin{figure*}[htbp]
	\vspace{-0.2cm}
	\centering
	\small
	\begin{tabular}{cc}
		\includegraphics[width=0.25\textwidth]{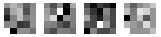}& \includegraphics[width=0.3\textwidth]{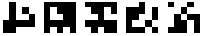}\\
		(a) The full-stack filters. & (b) The shared masks.\\
		\includegraphics[width=0.4\textwidth]{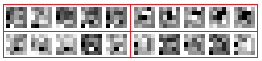}& \includegraphics[width=0.4\textwidth]{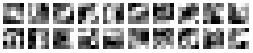}\\
		(c) The generated sub-filters. & (d) The convolution filters in original LeNet.\\
	\end{tabular}
	\vspace{-0.2cm}
	\caption{Visualization of convolution filters in MVNet-A and LeNet learned on MNIST, Filters in the same box are generated from the same full-stack filter.} 
	\label{fig:vis} 
	\vspace{-1.0em}
\end{figure*}

\subsubsection{Filter visualization}
To further investigate the effect of the proposed MVNet, we visualize the filters in the first layer in Fig. \ref{fig:vis} (more details can be found in supplementary materials). For an easy comparison, the original filters in LeNet are also shown in Fig. \ref{fig:vis} (d). We can see that the 5 masks are extremely different from each other, which then leads to distinct sub-filters. The 4 full-stack filters are also very diverse, and by combining the masks and the full-stack filters, the generated sub-filters would have various  structures. It is interesting that some generated sub-filters and the original filters in LeNet share similar patterns, such as skew edge and vertical texture. But other sub-filters may look different, as they are specially configured for the subsequent layers. Sub-filters generated from the same full-stack filter demonstrate different patterns. Hence, the learned binary masks and full-stack filters could cooperate to extract various patterns from the input feature map.

\begin{table}[htbp]
	\vspace{-0.2cm}
	\centering
	\caption{The performance of the proposed full-stack filters on CIFAR-10.}\label{tab:cifar10}
	\small
	\begin{tabular}{c|c||c|c|c}
		\hline
		{Method}   & {Architecture} & Memory (MB) & \#MUL (M) & ACC (\%) \\
		\hline\hline
		Original &\multirow{4}{*}{VGG-16} &  56.1 & 314.2 & 93.5 \\
		MVNet-A ($s$=4) &  & 14.1 & 78.5 & 92.9 \\
		MVNet-B ($s$=4) &  & 16.0 & 78.5 & 93.4 \\  
		MVNet-B ($s$=64) & & 2.7 & 4.9 & 93.1 \\\hline
	\end{tabular}
\vspace{-1.0em}
\end{table}

\subsection{Experiments on CIFAR-10}
The CIFAR-10 dataset consists of 60,000 $32\times32$ color images in 10 classes, with 6,000 images per class. There are 50,000 training images and 10,000 test images. 
VGG-16 \cite{Simonyan15} is originally designed for ImageNet classification, and its variant~\cite{cifar-vgg,retinking-pruning} which is widely used for CIFAR-10 is used here as the baseline model for CIFAR-10. All convolutional layers are replaced with the proposed convolution, and the compression performance is listed in Table \ref{tab:cifar10}. Under the setting $s=4$, MVNet with separate masks could obtain comparable performance and reduce $70\%$ parameters. Sharing masks in MVNet-A further reduces the memory usage but decreases the accuracy slightly.

To investigate the performance under an extreme compression setting, we set $s=64$ which implies compressing VGG-16 with about a $21\times$ ratio. The accuracy still reaches $93.1\%$ surprisingly, with only $0.4\%$ lower than that of the original model. Our method can achieve extremely large compression ratios and obtain very subtle performance decrease.

\subsection{Experiments on ImageNet}
Next we employ the proposed convolution for CNN compression on a large-scale dataset, namely, ImageNet ILSVRC 2012 dataset, which contains over $1.2M$ training images and $50k$ validation images belonging to $1,000$ classes. We evaluate the compression performance of our method on two widely used conventional models: VGG-16 \cite{Simonyan15} and ResNet-50 \cite{he2016deep}. In order to valid the compression performance, we replace all the conventional convolution layers with our convolution, except the last fully connected (FC) layer for classification (all the fully connected layers except the last one in VGG-16 are regarded as convolution layers). Here all training settings such as weight decay, learning rate and data augmentation strategy follow the settings in \cite{he2016deep} for fair comparisons. 

\begin{table*}[htb]
	\vspace{-0.3cm}
	\centering
	\renewcommand{\arraystretch}{1.05} 
	\caption{The performance of the proposed MVNet on ImageNet. Mem means memory usage of the entire model, Top1err and Top5err means top-1 and top-5 error rate, respectively.}\label{tab:imagenet}
	\small
	\begin{tabular}{c|c||c|c|c|c|c}
		\hline
		\multirow{2}{*}{Method} & \multirow{2}{*}{Architecture} & \#param & Mem & \#MUL & Top1err  & Top5err\\
		& & (32-bit) & (MB) & (B) & (\%)  & (\%)\\
		\hline\hline
		Original \cite{Simonyan15} & \multirow{5}{*}{VGG-16} & $1.38\times10^8$ & 573 & 15.4 & 28.5 & 9.9 \\
		ThiNet-Conv \cite{luo2017thinet} & & $1.31\times10^7$ & 500 & 9.6 & 30.5 & 10.2 \\
		BN low-rank \cite{tai2016convolutional} & & $0.50\times10^8$ & 191 & 5.0 & - & 9.7 \\
		MVNet-A ($s$=4)  & & $0.38\times10^8$ & 144 & 3.9 & 30.7 & 10.7 \\
		MVNet-B ($s$=4)  & & $0.42\times10^8$ & 160 & 3.9 & 28.9 & 9.7 \\
		\hline
		Original \cite{he2016deep} & \multirow{8}{*}{ResNet-50} &  $2.56\times10^7$ & 97.3 & 4.1 & 24.7 & 7.8 \\
		Versatile-v2 \cite{Versatile} & & $1.10\times10^7$ & 41.7 & 3.0 & 25.5 & 8.2 \\ 
		Slimmable 0.5$\times$ \cite{slimmable} & & $0.69\times10^7$ & 26.3 & 1.1 & 27.9 & - \\
		ShiftResNet \cite{Wu_2018_CVPR} & & $0.60\times10^7$ & 22.9 & - & 29.4 & 10.1 \\
		MVNet-A ($s$=4) & & $0.80\times10^7$ & 30.4 & 1.0 & 26.8 & 8.9 \\
		MVNet-B ($s$=4) & & $0.87\times10^7$ & 33.2 & 1.0 & 25.5 & 8.0 \\
		\cline{1-1}\cline{3-7}
		MVNet-B ($s$=32) & & $0.36\times10^7$ & 13.6 & 0.13 & 26.5 & 8.7 \\
		\hline
	\end{tabular}
	\vspace{-0.5em}
\end{table*}

After training on the ImageNet dataset, MVNets compress VGG-16 by $3-4\times$ and maintain the accuracies well. Similar to the observations in MNIST experiments, MVNet-B performs better than MVNet-A. MVNet-B obtains a $9.7\%$ top-5 error, which is even better than the original VGG-16. 

ResNet has introduced shortcut operations and various receptive fields to utilize filters, so it has less redundancy and it is challenging for further compression. The comparison between MVNet and original ResNet-50 is listed in Table \ref{tab:imagenet}. When setting $s=4$, MVNet-A has an accuracy drop of about 2 points, while MVNet-B has a better accuracy, achieving an $8.0\%$ top-5 error which is comparable with that of the baseline model. It is surprising to see that when we further increase $s$ to 32 in all the convolutional layers in MVNet-B, the top-5 accuracy of MVNet only decreases $0.9\%$ from that of the original model. In this case, the model size is reduced to 13.6 MB which is suitable for the mainstream cache capacity in mobile devices. In addition, MVNet can reduce the number of MUL by about $32\times$ as described in Section \ref{sec:mul}. Detailed statistics of VGG-16 and ResNet-50 can be found in the supplementary materials.

\subsubsection{Comparison with Existing Efficient CNNs}
We next compare the proposed MVNets with state-of-the-art portable CNN architectures, such as MobileNet \cite{mobilev2}, ShuffleNet \cite{zhang2018shufflenet,shufflev2}, \etc. We base our MVNets on MobileNet-v2 by replacing $1\times1$ convolutions with the proposed convolution operations. Table \ref{tab:portable} summarizes the statistics of these architectures, including their memory usages and recognition accuracies. Unlike VGG-16 or ResNet, MobileNet already has more compact structure and less redundancy, thus further compression seems to be more challenging. Here we employ MVNet-B with separate masks for promoting diversity in filters. By exploiting the proposed method on MobileNet-v2 with $s=4$, the performance of MVNet has a small decrease but is still better than that of other portable models with a smaller model size (7.5 MB) and significantly fewer multiplication operations (93 M).


\begin{table}[htb]
	\vspace{-0.5em}
	\centering
	\renewcommand{\arraystretch}{1.05} 
	\caption{Comparison with existing efficient CNNs on ImageNet.}\label{tab:portable}
	\small
	\begin{tabular}{c||c|c|c|c}
		\hline
		{Method} & \#param (32-bit) & Mem (MB) & \#MUL (M) & Top1err (\%) \\
		\hline\hline
		ShuffleNet-v1 1.5$\times$ (g=3)~\cite{zhang2018shufflenet}   & $3.4\times 10^6$& 13.0   & 292     & 31.0   \\
		ChannelNet-v2~\cite{gao2018channelnets} & $2.7\times 10^6$ & 10.3 & 361 & 30.5 \\
		0.75 MobileNet-v2~\cite{mobilev2} & $2.6\times 10^6$& 10.0 & 209 & 30.2  \\
		ShuffleNet-v2 1$\times$~\cite{shufflev2} & $2.3\times 10^6$& 8.7 & 146 & 30.6  \\
		AS-ResNet-w50~\cite{active-shift} & $2.0\times 10^6$ & 7.5 & 404 & 30.1 \\
		\hline
		1.0 MobileNet-v2~\cite{mobilev2} & $3.5\times 10^6$ & 13.2 & 300 & 28.2 \\
		MVNet-B ($s$=4, 1.0 MobileNet-v2) & $2.0\times 10^6$ & 7.5 & 93 & 29.7 \\
		\hline
	\end{tabular}
	\vspace{-0.5em}
\end{table}

\subsection{Experiments on Object Detection}
To investigate the generalization ability of the proposed MVNet, we further evaluate it on the MS COCO object detection task \cite{lin2014microsoft}. We adopt Faster-RCNN \cite{renNIPS15fasterrcnn} with 600 input resolution as the detection framework and use ResNet-50 as a backbone architecture. The models are trained on the COCO train+val dataset excluding 5,000 minival images and we conduct test on the minival set. From the results in Table~\ref{tab:coco}, MVNet can compress ResNet-50 by $3.5\times$ and obtain comparable performance. Some detection examples are listed in supplementary materials.

\begin{table}[htbp]
	\centering
	\renewcommand{\arraystretch}{1.05} 
	\caption{The object detection results on MS COCO. Conv Mem means memory usage of the convolutional layers. mAP is reported with COCO primary challenge metric (AP at IoU=0.50:0.05:0.95)} \label{tab:coco}
	\small\vspace{0.1cm}
	\begin{tabular}{c||c|c|c}
		\hline
		{Method} & Conv Mem (MB) & \#MUL (B) & mAP (\%) \\
		\hline\hline
		Original  & 89.7 & 50.4 & 33.1 \\
		MVNet-B ($s$=4)  &  25.4  & 12.6 & 32.0 \\
		\hline
	\end{tabular}
	\vspace{-0.3cm}
\end{table}

\section{Conclusion}
In this paper, we propose a novel convolution operation with full-stack filters and auxiliary masks to reduce redundancy in convolution filters for building minimum viable CNNs, \ie MVNets. A series of full-stack filters with additional binary masks which requiring negligible memory resource in the proposed convolution method can generate much more inherited sub-filters and achieve approximate performance to the original convolution. In addition, a specific optimization algorithm is designed for the optimization of full-stack filters and masks. Experimental results on the benchmark dataset show that the proposed method can efficiently reduce the memory usage and multiplication computation, leading to minimal viable models with satisfactory performance.

\clearpage\clearpage

\bibliography{MVNet_ref}
\bibliographystyle{ieee}

\end{document}